\documentclass[12pt]{article}

\usepackage[utf8]{inputenc} 
\usepackage[T1]{fontenc}    
\usepackage{url}            
\usepackage{booktabs}       
\usepackage{amsfonts}       
\usepackage{nicefrac}       
\usepackage{microtype}      
\usepackage{xcolor}         
\usepackage{soul}
\usepackage{booktabs}
\usepackage{hyperref}       
\hypersetup{
    colorlinks=true,
    citecolor=magenta,
    linkcolor=blue,
    filecolor=blue,      
    urlcolor=cyan,
    pdftitle={FEAR},
    }

\usepackage{caption}
\usepackage{subcaption}
\usepackage{wrapfig}
\usepackage{algorithm}
\usepackage{algorithmic}
\usepackage{makecell}
\usepackage{natbib}

\usepackage{titling}
\predate{}
\postdate{}

\usepackage{graphicx}
\graphicspath{ {./images/} }

\newcommand{\methodname}{\texttt{FEAR} }
\newcommand{\jacobcov}{\texttt{jacob\_cov} }
\newcommand{\snip}{\texttt{snip} }
\newcommand{\grasp}{\texttt{grasp} }
\newcommand{\synflow}[1][]{\texttt{synflow#1} }
\newcommand{\synflowbn}{\texttt{synflow\_bn} }
\newcommand{\fisher}{\texttt{fisher} }
\newcommand{\vote}{\texttt{vote} }
\newcommand{\gradnorm}{\texttt{grad\_norm} }
\newcommand{\shortreg}{\texttt{shortreg} }
\newcommand{\synthetic}{Synthetic CIFAR10 }

\title{\methodname: A Simple Lightweight Method to Rank Architectures}

\author{%
  Debadeepta Dey \\
  Microsoft Research \\
  \texttt{dedey@microsoft.com} \\
  \and
  Shital Shah \\
  Microsoft Research \\
  \texttt{shitals@microsoft.com} \\
  \and
  S\'ebastien Bubeck \\
  Microsoft Research \\
  \texttt{sebubeck@microsoft.com} \\
}

\date{} 

\begin{document}

\maketitle

\begin{abstract}
  The fundamental problem in Neural Architecture Search (NAS) is to efficiently 
  find high-performing architectures from a given search space. We propose a simple
  but powerful method which we call \methodname, for ranking architectures in any search space. \methodname
  leverages the viewpoint that neural networks are powerful non-linear feature extractors.
  First, we train different architectures in the search space to the same training or validation
  error. Then, we compare the usefulness of the features extracted by each architecture. We do so
  with a quick training keeping most of the architecture frozen. This gives fast estimates
  of the relative performance. We validate \methodname on Natsbench topology search space
  on three different datasets against competing baselines and show strong ranking correlation especially
  compared to recently proposed zero-cost methods. \methodname particularly excels
  at ranking high-performance architectures in the search space. When used in the inner loop of 
  discrete search algorithms like random search, \methodname can cut down the search time by 
  $\approx2.4$X without losing accuracy. We additionally empirically study very recently 
  proposed zero-cost measures for ranking and find that they breakdown in ranking performance as training proceeds 
  and also that data-agnostic ranking scores which ignore the dataset do not generalize 
  across dissimilar datasets.
\end{abstract}

\section{Introduction}
\label{introduction}
Neural Architecture Search (NAS) \citep{elsken2019neural,ren2021comprehensive} is a sub-field of
automatic machine learning (AutoML) \citep{heAutomlSurvey,automl_book} where the aim is for algorithms to search for 
high-performing architectures instead of humans manually trying out many possibilities. Given the 
rapid advances in differentiable operator types and usual application-dependent open questions of number of 
layers, channels, input resolution, etc this leads to an explosion of combinatorial choices in architecture
design space. Indeed, even ubiquitously used search spaces like the DARTS \citep{liu2018darts} search space
contain approximately $10^{18}$ architectures \citep{siems2020nasbench301}. The fundamental problem in NAS is to 
search this combinatorially exploding space of architectures as efficiently as possible.

Current NAS methods conduct this search in one of mainly two different ways: 1. \textbf{Discrete methods}
use methods that sample architectures from a search space, estimate the final performance of the architectures 
by training them partially and then update the sampling rule based on that estimate \citep{nas_via_rl_2017,liu2021survey}. These
methods can be notoriously computationally expensive due to the cost of training each sampled architecture individually \citep{whiteBananas,white2020local}.
2. \textbf{One-shot methods} on the other hand train a single large graph (colloquially known as `supergraph')
that contains all possible architectures in one single graph by sharing weights amongst common edges in architectures. 
Even though weight-sharing has many issues like optimization gap between training and evaluation and shallow sub-graphs training faster \citep{xie2020weightsharing,wang2021rethinking,understandingCellNas,Zela2020Understanding}, by training a single graph the cost of gradient computation via backpropagation can be amortized over exponentially many sub-graphs. Popular approaches include bilevel optimization \citep{liu2018darts}, single level optimization approaches \citep{data_differentiable_approx} utilizing the Gumbel-Softmax trick \citep{jangCategorical,maddisonConcrete} or even direct optimization without architecture weights \citep{wang2021rethinking}. Note that for both with and without weight-sharing one can use various classes of techniques. For example RL-based \citep{phamEnas}, evolutionary search \citep{liu2021survey}, random search \citep{liRandomSearch}, local search \citep{white2020local} and variants of Bayesian optimization \citep{whiteBananas}.

\begin{figure}
  \centering
  \includegraphics[width=\textwidth]{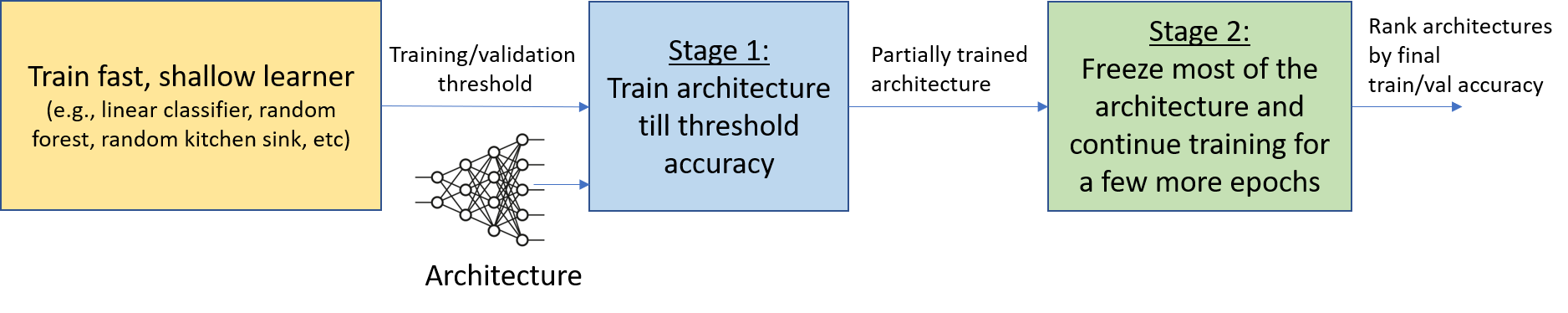}
  \caption{Overview of \methodname which first trains a fast but shallow learner to get a reasonable training/validation error threshold and then trains the architecture in a two-stage procedure. In the first stage the architecture is regularly trained until it achieves threshold accuracy. In the second stage most of the partially trained architecture from the first stage is frozen and training continues for a few more epochs. All candidate architectures can then be ranked by final training or validation accuracy obtained via this two stage procedure.}
  \label{fig:schematic}
\end{figure}

We focus on the evaluation phase of discrete methods where it is often an ad-hoc choice on how and for how
long to evaluate each sampled architecture for a given dataset. As \citet{liRandomSearch} note, the partial training phase can often be critical in deciding how well a method performs. Crucially, they note that using partial training or validation error after many epochs of 
training can still not get the correct final rank of architectures. We propose a simple but powerful architecture 
ranking methodology that enables fast architecture ranking by leveraging the fact that neural networks are powerful 
feature extractors and the power of an architecture is dependent on how effective it is at extracting useful features from the inputs for the given task. We term this `recipe' (see Figure \ref{fig:schematic} for an overview) for ranking architectures \methodname (FEATure-extraction Ranking). At a high level, first trains any architecture regularly till the architecture attains a specified training or validation accuracy threshold in the first stage. In the second stage, most of the architecture is frozen and the final few layers continue training for a bit longer. Architectures can be relatively ranked via their training or validation error after these two stages. We leverage insights in training dynamics of neural networks \citep{kornblith2019similarity,svccaRaghu} that show that early layers train very fast and become stable within a few epochs but later layers spend most of their time learning to make decisions from the features extracted by the earlier layers. A crucial aspect of our procedure is that we train architectures up to a specified threshold accuracy instead of committing apriori to fixed number of epochs. This exposes first order dynamics in neural network training where invariably (empirically) weak architectures take much longer to reach threshold accuracy as opposed to stronger architectures (see Figure \ref{fig:three_cond_time}). Such first order training dynamics can be exploited in any sampling-based discrete NAS algorithm. \methodname is discussed at length in Section \ref{approach}.

We have two main contributions in this work:
\begin{itemize}
    \item We propose a simple fast architecture evaluation method named \methodname and validate it on a variety of datasets on the Natsbench topological space benchmark against competing baselines.\footnote{Reproducible implementation of all experiments is available at \url{https://github.com/microsoft/archai/tree/fear_ranking} \citep{archai} under MIT license. } 
    \item We also empirically find that a number of very recently proposed lightweight ranking measures \citep{abdelfattah2021zerocost,mellor2021neural} degrade in ranking performance as network training progresses and that data-agnostic ranking measures don't generalize across datasets. The performance of an architecture is a function of \emph{both the topology and the dataset} (in addition to the training pipeline).  
\end{itemize}

\section{Related Work}
\label{related_work}

NAS research has received a lot of attention recently. We refer the reader to continuously updated excellent surveys  
\citep{elsken2019neural,ren2021comprehensive} and a near-exhaustive list of papers at \citep{naslit} for an overview of the field. Here we discuss the works that are directly relevant to fast evaluation and ranking of architectures.

\paragraph{Architecture Performance Prediction:}\citet{baker2017accelerating} propose training regressors which take in the architecture,
training hyperparameters and the first few validation accuracies as features and try to predict the final validation 
accuracy. The proposed method has a ``burn-in'' phase where a sampled few architectures are first fully trained to 
gather data for training the regressor. This regressor is then utilized in the rest of the pipeline. 
Similarly \citet{rorabaugh2021peng4nn} propose fitting curves to the performance numbers of the first few training iterations 
and extrapolate to later accuracy values. \citet{whiteBananas} study neural network performance prediction in the context of Bayesian Optimization.
Orthogonally, \citet{white2020study} conduct an extensive study of architecture encodings for common NAS
algorithm subroutines including performance prediction which sheds light on the pros and cons of certain featurizations for
this task. \methodname is orthogonal to the above body of work on architecture performance prediction and in fact
can be used for further speeding up the performance prediction modules as one doesn't have to train the architectures
fully. 

\paragraph{Lightweight Architecture Evaluation:}\citet{zhou2020econas} search for combinations of reduced input image resolution,
fewer epochs, and number of stem channels to find computationally cheap proxies for evaluating architectures while keeping their
relative ranks the same. They find an optimal configuration of resolution, epochs and number of channels on a bag of $50$ models.
They term their method as EcoNAS. \citet{abdelfattah2021zerocost} note that the configuration found by EcoNAS suffers from degrading performance 
when evaluated on all $15625$ models in Nasbench-201 CIFAR10 dataset \citep{Dong2020NASbench201}. \citet{abdelfattah2021zerocost} conduct their own search and find
a different configuration that works better on Nasbench201 CIFAR10 dataset. They caution that such proxies clearly don't
work on different search spaces even when the dataset and task are the same and also the importance of measuring actual wall clock
run times as reduced flops often don't translate into actual time savings due to different ways of accessing memory.

\citet{cao2021efficient} propose a lightweight architecture evaluation method based on the viewpoint that neural networks are rich feature 
extractors which are utilized by the last linear classifier layer. They propose training an architecture for a few epochs and recording
the features per datapoint over a window of `k' last epochs. A linear classifier is trained on each of these feature histories to 
produce `k' linear classifiers. An ensemble of these `k' classifiers is used to predict the class membership of each datapoint. The average
error over all datapoints is used to rank architectures. This method is closest in spirit to \methodname but differs in a number of 
important ways. First of all maintaining feature histories for every datapoint is expensive in terms of memory. \methodname avoids
this step by simply freezing most of the architecture and continuing training (which is much cheaper since gradients have to be computed
only for a small part of the architecture). More critically \methodname \emph{does not apriori fix the number of epochs}. Instead
it trains the network in the first stage to a prespecified threshold training or validation accuracy and then freezes most of the 
network in the second stage. We describe in detail in Section \ref{approach} how to organically set this threshold in a 
task-dataset dependent manner. 

\paragraph{Trainingless Proxies:}\citet{mellor2021neural} propose a \emph{trainingless} method for ranking architectures based on the KL divergence
between an uncorrelated Gaussian distribution and the correlation matrix of local linear operators (for networks with ReLU activation
functions) associated with every input data point. If the correlation between such local linear maps is low
then the network should be able to model each data point well. Since this score can be computed with just a small sample of the 
dataset (typically a single minibatch), this takes negligible compute and time. In very recent work \citet{abdelfattah2021zerocost} thoroughly empirically 
evaluate this trainingless method which they term as \jacobcov along with an entire family of pruning-at-initialization
schemes which they convert to trainingless architecture ranking methods by simply summing up the saliency scores at each 
weight of the architecture. The particular methods they evaluate include \snip \citep{lee2018snip}, 
\grasp \citep{Wang2020grasp}, \synflow \citep{tanaka2020synflow} and \fisher \citep{Turner2020BlockSwapfisher} 
in addition to other natural baselines like \gradnorm which is
the sum of Euclidean norm of the gradients using a single minibatch of training data. On Nasbench-201, on all three datasets 
(CIFAR10, CIFAR100, ImageNet16-120) they find that \synflow score performed the best with relatively high rank
correlations with full training of architectures. \jacobcov was second best. A majority vote amongst
\synflow, \jacobcov and \snip termed as \vote performs the best. They also note that 
these trainingless methods don't work satisfactorily when evaluated on other search spaces like Nasbench-101 \citep{ying2019nasbench101}. 

Note that \methodname is \emph{not} a trainingless method and does use more computation than the trainingless proxies outlined above.
But we empirically show in Section \ref{experiments} that \methodname outperforms these proxy measures as well as the natural baselines of 
reduced number of training epochs. Furthermore in the process of experimentation we identify some curious properties of
trainingless proxies such as degradation in performance as the network trains more which is counterintuitive and also 
the curious phenomenon of \synflow[-based] ranking (which is a data-agnostic scoring mechanism) in particular not generalizing across
datasets. This supports the intuition that architecture performance is not an intrinsic property of just its topology (and 
training procedure) but crucially also dependent on the task and dataset at hand. This has been empirically validated 
by the very recent work of \citet{tuggener2021optimize} who show that architectures which perform well on ImageNet \citep{imagenet} do not necessarily
perform as well on other datasets. In fact on some datasets their ranks are negatively correlated with ImageNet ranks. This 
further suggests that a data-agnostic scoring mechanism such as \synflow may not work well at ranking architectures.

\section{Approach}
\label{approach}


Here we describe \methodname in more detail. Figure \ref{fig:schematic} shows a high level schematic of the 
approach. 

\paragraph{Finding training accuracy threshold:}\methodname first trains a fast but shallow learner on 
the dataset and task of choice to learn a training or validation accuracy threshold. For example, for the task of image 
classification one can use a number of fast shallow learners like random forest, linear classifier,
with handcrafted visual features such as Histogram-of-Gradients (HoG) \citep{dalal2005histograms} or random features such as
random kitchen sink \citep{rahimi2008weighted}\footnote{See \citep{localdesc2005} for a survey of handcrafted 
visual features.}. We emphasize that the role of this threshold is to be both non-trivial yet not too difficult to beat
with a neural network architecture. We explain the intuition behind this choice below.

\paragraph{Stage 1: Regular training till threshold accuracy:} \methodname then trains the candidate
architecture till it achieves this threshold accuracy.

\paragraph{Stage 2: Using architecture as feature extractor:} 
\methodname then freezes most of the layers of the architecture other than the last few layers and trains it 
for a few more steps. This \emph{freezing} has the advantage of being several times faster per step 
than training the entire architecture as gradients don't have to be computed for most of the layers. This
stage essentially treats the network as a feature extractor and trains a relatively shallow network 
utilizing these features for a few more epochs. A pool of candidate architectures are then
ranked by their final training or validation accuracies on the dataset under consideration.
Intuitively, \methodname ranks architectures on their ability to extract useful features from inputs.

A question that may naturally arise is by cutting off training of most of the layers at a relatively
early stage of training, are we not hurting the architecture's ability to potentially distinguish itself
at feature extraction? \citet{svccaRaghu} and \citet{kornblith2019similarity} dive deep into the training dynamics of neural networks
and show that networks train `bottom-up' where the bottom layers (near the input) train quite fast early-on 
in training and become stable. As training progresses these bottom layers rarely change their representation
and mostly the top layers change to learn the decision-making rules using the bottom layers as rich feature
extractors. \methodname leverages this insight by first training all layers for a few epochs and then 
freezing most of the layers but the last few layers.

\begin{figure}
     \centering
     \begin{subfigure}[b]{0.3\textwidth}
         \centering
         \includegraphics[width=\textwidth]{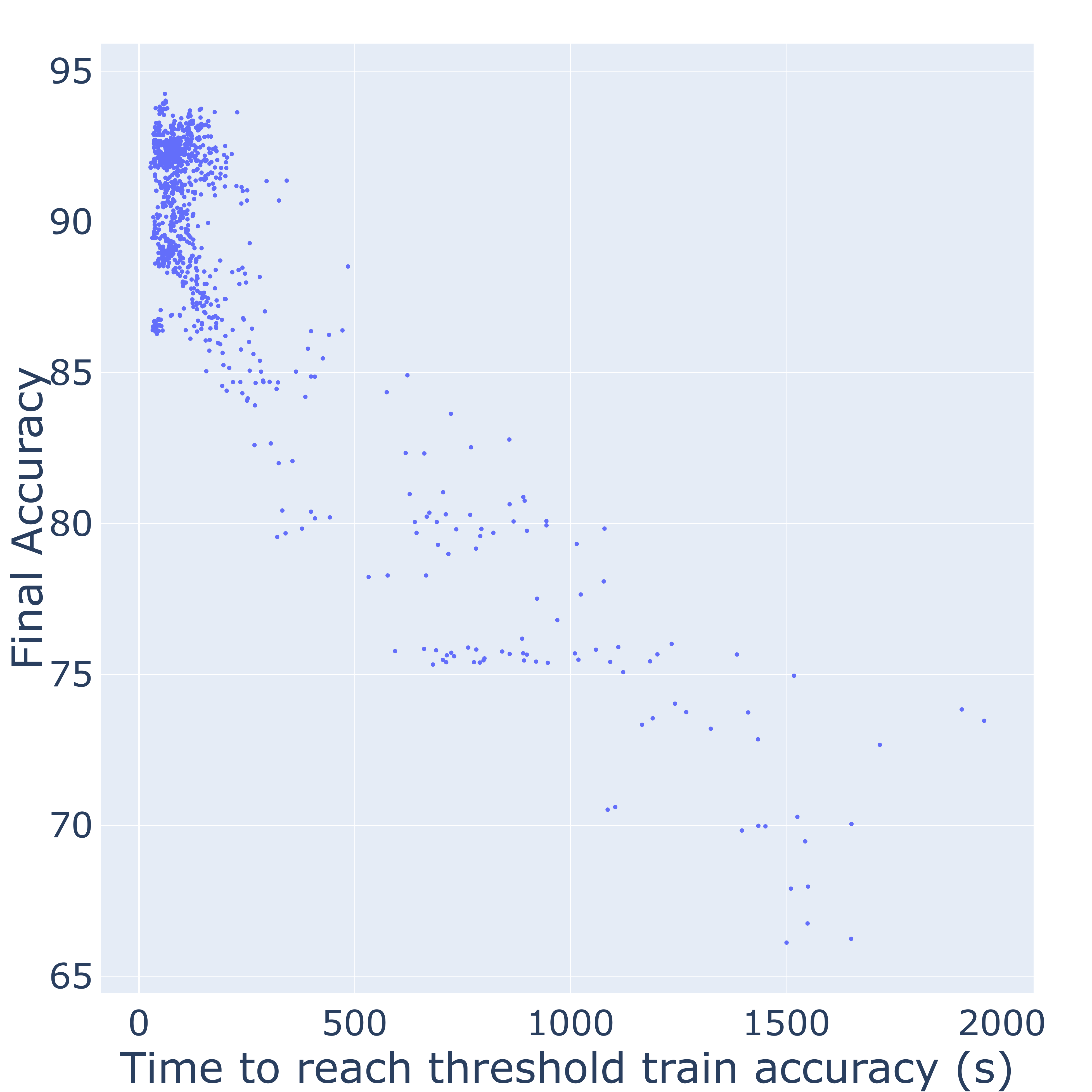}
         \caption{CIFAR 10}
         \label{fig:cond_time_cifar10}
     \end{subfigure}
     \hfill
     \begin{subfigure}[b]{0.3\textwidth}
         \centering
         \includegraphics[width=\textwidth]{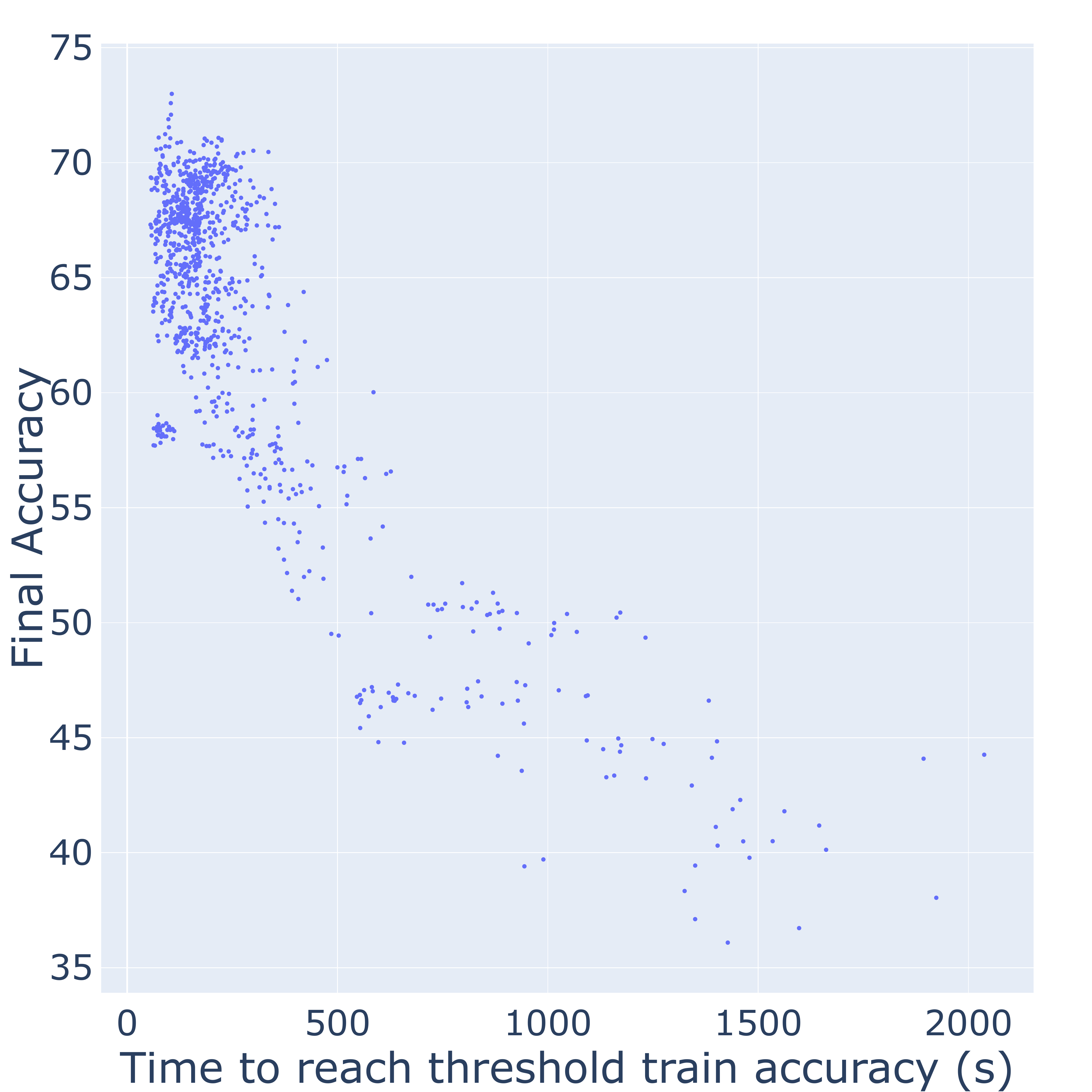}
         \caption{CIFAR 100}
         \label{fig:cond_time_cifar100}
     \end{subfigure}
     \hfill
     \begin{subfigure}[b]{0.3\textwidth}
         \centering
         \includegraphics[width=\textwidth]{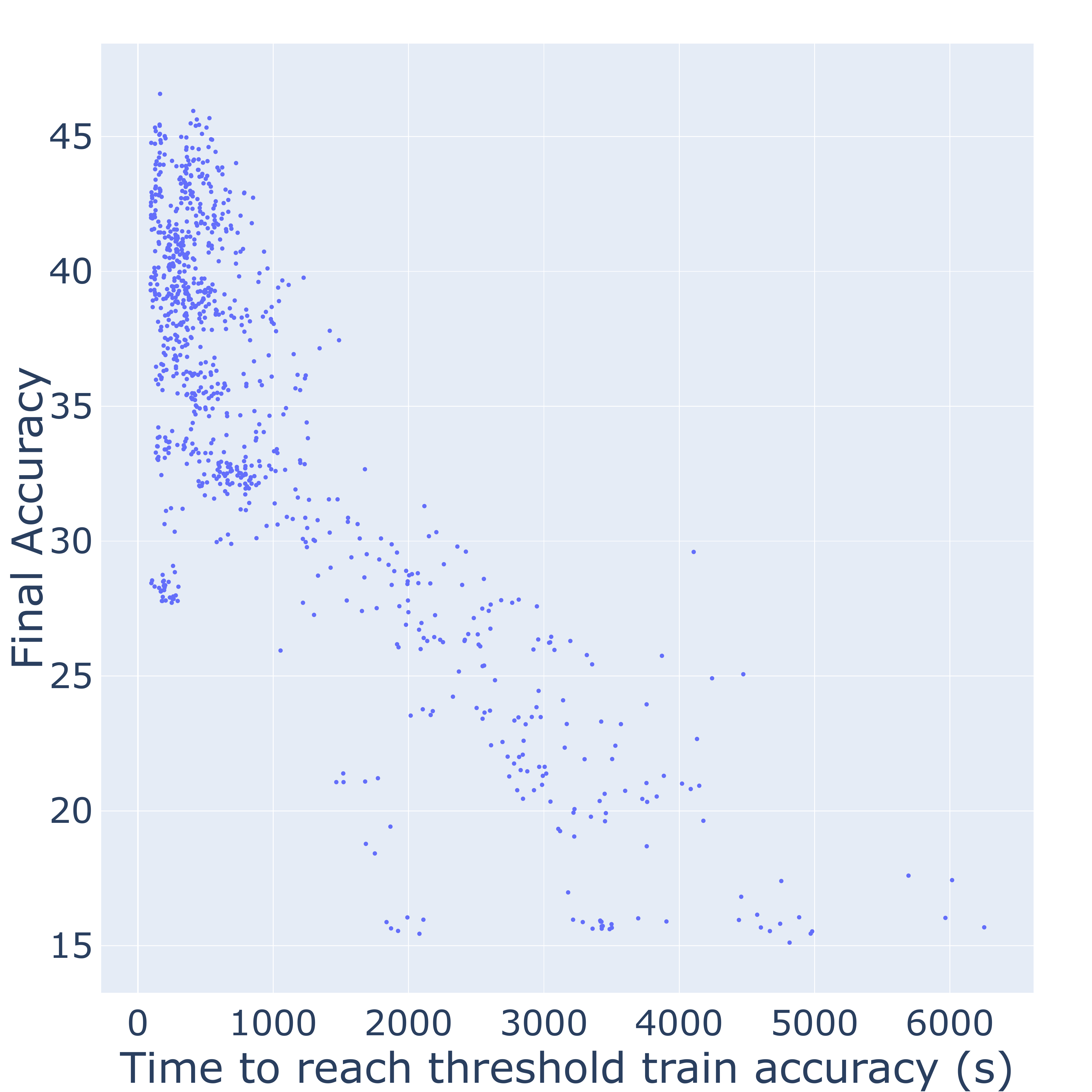}
         \caption{ImageNet16-120}
         \label{fig:cond_time_imagenet16-120}
     \end{subfigure}
        \caption{Time to reach threshold training accuracy (x-axis) vs. final test accuracy on $1000$ uniformly 
        sampled architectures on Natsbench. They show a clear relationship where ultimately worse performing architectures
        take longer time to reach threshold accuracy than stronger ones.}
        \label{fig:three_cond_time}
\end{figure}

\paragraph{Role of training till threshold accuracy:} We would like to re-emphasize that \methodname does not fix the number of epochs apriori. Instead it trains the architecture until a threshold accuracy has been reached.
This has a number of advantages. First, it makes architectures \emph{comparable} to each other and makes sure that every architecture gets ample time to learn the best features it can for the task. Secondly, it exposes first-order dynamics of training i.e. weak architectures emperically take longer time to reach the same training or validation error compared to the stronger architectures. See Figure \ref{fig:three_cond_time} where we plot for three datasets the time
taken by $1000$ architectures to reach a threshold accuracy (x-axis) against the final test error (y-axis). Invariably, we find that architectures which go on to attain good final test accuracy achieve threshold training accuracy much faster than weaker ones. This alone is not enough for good ranking as some weaker architectures can achieve the threshold accuracy fast as well and hence ranking by training a bit longer to resolve the power of features is necessary.
Also note that there are no architectures that train slowly but go on to achieve good final test accuracy after full training (`late-bloomers', these would have been on the upper-right hand part of the plots).
This effect can be utilized to early-stop evaluation of candidates that take much longer than the fastest architecture encountered to reach threshold accuracy. This is in fact crucial to obtain large speedup when combining our ranking method with standard discrete neural architecture search
techniques.

\paragraph{Motivation:} The motivation for our method comes from the emerging theoretical understanding of gradient learning on neural networks. In a number of works (e.g., \citep{hu2020surprising,chizat2020implicit,nakkiran2019sgd,allenzhu2021backward}) it has been observed that at first training happens in the so-called ``neural tangent kernel'' (NTK) \citep{ntk2018} regime, where the network basically uses its initialization as a sort of kernel embedding and performs a kernel regression. Our key hypothesis is that this phase should stop a bit before reaching the fixed threshold accuracy, since this threshold has been obtained with a good kernel method (or something slightly more powerful like a random forest model). 
To put it differently, in our method, when we stop the training after reaching the fixed threshold accuracy, it should be that the network has already escaped the NTK regime and is currently actively training the features (second phase of learning). Our second (empirical) hypothesis is that the quality of the features learned in the early part of this second phase is predictive of the final quality of the network. 
We measure the quality of the learned features via the freezing technique, whose extreme case is to only continue training the final layer (i.e., train a linear model on top of the current embedding).

\section{Experiments}
\label{experiments}

\paragraph{Search space:} We uniformly randomly sample $1000$ architectures from the $15625$ architectures of the
topology search space of the Natsbench \citep{natsbench} benchmark\footnote{Note that Natsbench topology search space is 
the same as Nasbench-201 \citep{Dong2020NASbench201}.} and 
hold them constant for all following experiments. The topology search space is similar to that used in DARTS \citep{liu2018darts}. All cells have the same topology. 
Edges are different operator choices (the set of operators are \texttt{zeroize}, \texttt{skip connection}, $1\times1$ \texttt{convolution}, $3\times3$
\texttt{convolution} and $3\times3$ \texttt{average pooling}). Nodes are tensors. Each cell has $4$ nodes. There are three stages in
the outer macro skeleton with a downsampling operation after each stage. Each cell is repeated $5$ times in each stage. 
There are respectively $16$, $32$ and $64$ channels in each stage. See Figure 1 in \citep{natsbench} for a visualization of the 
search space. Natsbench topology search space has trained each architecture on CIFAR10, CIFAR100 \citep{krizhevsky2014cifar}
and ImageNet16-120 \citep{downsampledImagenet} image classification datasets.

\subsection{Performance criteria}
\paragraph{Spearman's rank correlation vs. evaluation wall-clock time:} We report performance of \methodname and baselines by first binning architectures into several buckets of increasing size. For example \texttt{Top $10\%$} in Figure \ref{fig:cifar10_natsbench} shows the average wall-clock time taken by any method (x-axis) vs. Spearman's rank correlation \citep{spearman}\footnote{Spearman's 
rank correlation is between $(-1, 1)$ with $1$ implying perfect correlation and $-1$ anti-correlation of the \emph{ranks} of candidate architectures.} of the method with 
the groundtruth rank of architectures (by test accuracy) after full final training procedure over the top $10\%$ of architectures. Similarly
the bin of \texttt{Top $20\%$} architectures includes top $20\%$ of candidates and so on. We break-up the performance of methods over such cumulative bins to highlight how methods perform in discriminating amongst high-performing candidates
and not just over the entire population. It is crucial for any reduced-computation proxy ranking method to hone-in
on good ones and not just the entire population.

\paragraph{Percentage overlap with groundtruth ranking vs. evaluation wall-clock time:} While Spearman's rank correlation over cumulative bins of candidates by groundtruth performance is useful for showing the ability of methods to 
discriminate amongst the top $x\%$ of architectures, it is also important to evaluate what percentage of architectures are common between the top $x\%$ of groundtruth architectures and the architectures that are ranked by a method. When ranking the entire population of candidates, this metric evaluates if the high ranking architectures in groundtruth are also highly ranked by a method (and vice-versa). By definition this is a quantity between $(0, 1)$.  

\subsection{Baselines}

\paragraph{Regular training with reduced epochs: \shortreg} The most natural baseline is to compare rank correlation of \methodname against reduced epochs of training. Most NAS methods use a reduced number of training epochs \citep{liRandomSearch} in the inner loop to decide the relative ranks of architectures as a proxy for final performance after undergoing the complete training procedure. We term this reduced training proxy as \texttt{shortreg}. We show that \methodname consistently outperforms the pareto-frontier of wall-clock time vs. Spearman rank correlation and the ratio of common architectures over cumulative bins of candidate architectures by groundtruth (test accuracy).  

\paragraph{Zero-cost Proxies:} As detailed in Section \ref{related_work}, in very recent work \citep{abdelfattah2021zerocost} a number of nearly negligible cost proxies for ranking architectures are presented. We evaluate these zero-cost 
proxies for ranking and observe a number of mysterious phenomena where such measures 
break-down across datasets or as networks are trained. 
This shows that such measures while exhibiting reasonable prima-facie performance on NAS benchmarks are not generalizing across datasets. We detail our investigation 
in Section \ref{deep_dive_zerocost}.

\paragraph{Reduced Resolution Proxies:} While reduced resolution proxies as proposed in \citep{zhou2020econas}
and \citep{abdelfattah2021zerocost} can significantly speed up architecture evaluation, note that they are orthogonal 
to our approach as they equally speed-up both \shortreg and \methodname. For use in production pipelines 
they should be combined with \methodname to get even more speedup.

\paragraph{Training procedure hyperparameters and hardware:} In all our experiments we use the same
hyperparameter settings as Natsbench, specifically cosine learning rate 
schedule with starting learning rate of $0.1$, minimum learning rate of $0.0$, SGD optimizer, 
decay of $0.0005$ for both regular weights and batch-norm weights, momentum $0.9$ and Nesterov enabled.
For \shortreg baseline we run experiments with varying number of epochs and batch sizes $256, 512, 1024, 2048$
to find a pareto-frontier of wall-clock time vs. Spearman's correlation and common ratio. We especially
investigate varying the batch size since that can have a large effect on wall-clock time.
When architectures are evaluated by \methodname, for the second phase we froze the network
up to \texttt{cell13} out of $16$ total cells for all architectures in the search space. 
This usually corresponds to $\approx53\%$ of parameters.
All experiments were conducted on Nvidia V100 GPUs with 16 GB of GPU memory.

\subsection{Main Results}
\label{main_results}

\begin{figure}[htbp!]
  \centering
  \includegraphics[width=\textwidth]{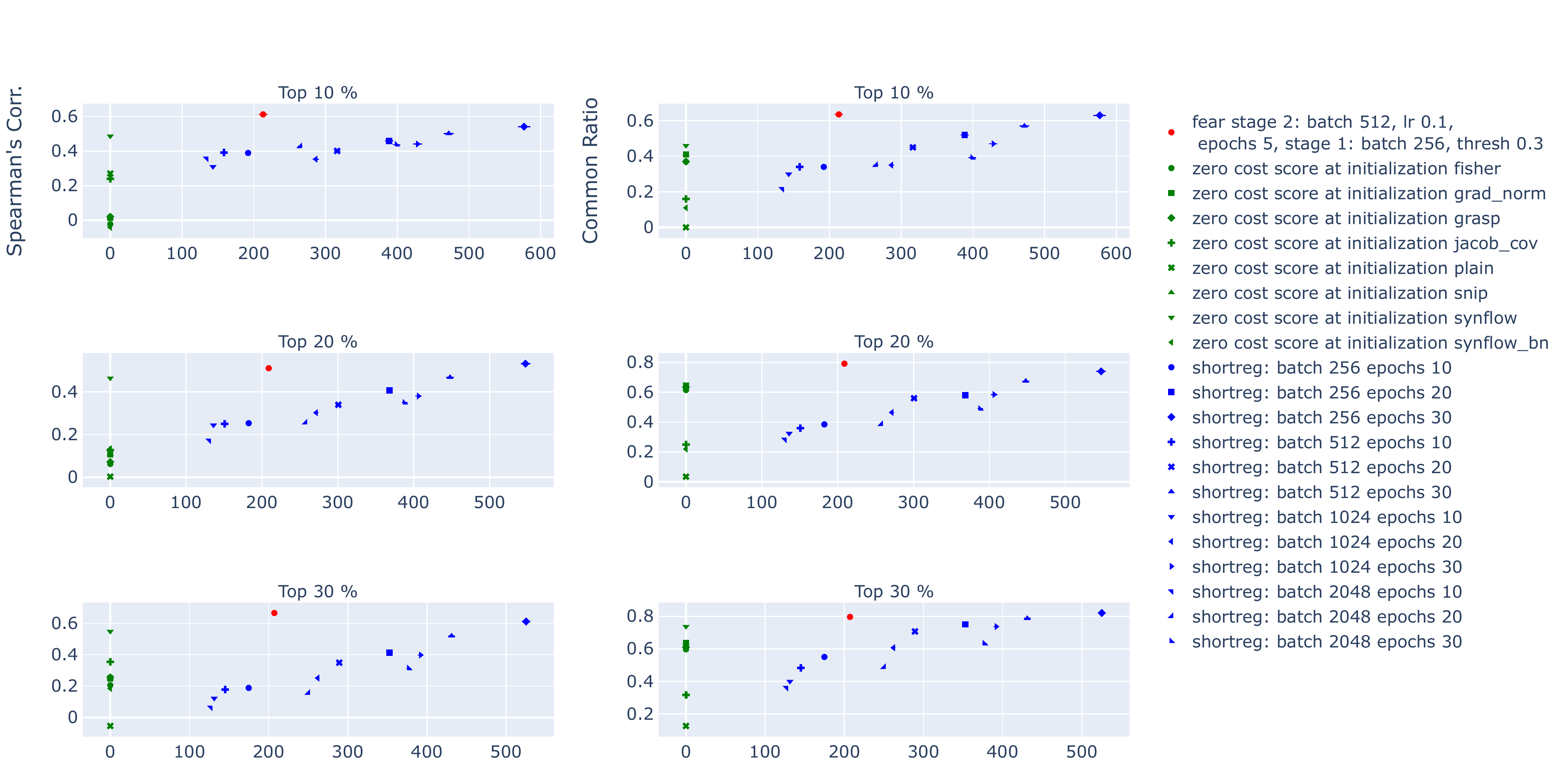}
  \caption{[Left] Average duration per architecture vs. Spearman's correlation and [Right] average duration per architecture
  vs. common ratio over the top $x\%$ of 
  the $1000$ architectures sampled from Natsbench topological search space on CIFAR100. We also show the various zero-cost measures
  from \citet{abdelfattah2021zerocost} in green. Recall that in Section \ref{deep_dive_zerocost} these measures will be shown to 
  be non-robust to change of task as well as degradation when the training progresses.}
  \label{fig:cifar100_natsbench_cropped}
\end{figure}

Figure \ref{fig:cifar100_natsbench_cropped} shows the ranking
efficiency of \methodname vs. \shortreg on the uniform random sample of $1000$ architectures on CIFAR100 up to the 
top $30\%$ of architectures. For elucidation purposes we give out detailed versions in Appendix \ref{app:detailed_ranking} 
(figures \ref{fig:cifar10_natsbench}, \ref{fig:cifar100_natsbench} and 
\ref{fig:imagenet16-120_natsbench}) on Natsbench CIFAR10, CIFAR100 and ImageNet16-120 respectively. We sort architectures in descending order of test accuracy, 
and bin them cumulatively into top $10\%$, $20\%$, $\ldots$ bins.
For each bin we report the two performance criteria detailed above of Spearman's rank correlation and the common ratio of architectures
in the global ranking. For CIFAR10 and CIFAR100 we find that \methodname consistently places above the pareto-frontier 
of \shortreg especially at higher ranked architectures. This means that \methodname is able to both discriminate better amongst
high-performing architectures (better Spearman's correlation) with shorter wall-clock time as well as achieve global ranking
which overlaps more with the groundtruth ranking of architectures using test accuracy. Note that as the bin increases to encompass the
entire set of $1000$ architectures (Top $100\%$), by construction, \methodname will start taking more time as low-performing 
architectures take more time to reach threshold accuracy (recall Figure \ref{fig:three_cond_time} and associated explanation)
and hence increase the total/average wall-clock time. In practice, this extra time for lower-performing architectures will 
not be paid since low-performing ones can be simply removed from consideration when they exceed some multiple of the fastest time so far by other architecures to achieve the same threshold accuracy. On ImageNet16-120 the gap between \methodname and \shortreg is not as big but nevertheless it 
doesn't degrade in performance below \shortreg and over the high-performance bins is marginally better.

Since figure \ref{fig:cifar100_natsbench_cropped} and corresponding detailed versions 
\ref{fig:cifar10_natsbench} \ref{fig:cifar100_natsbench} and \ref{fig:imagenet16-120_natsbench} contain dense
information, for ease of elucidation, in Table \ref{tab:three_spe_cr} for each bin we note \methodname and the nearest point 
on the pareto-frontier generated by \shortreg to rank them amongst each other using Spearman's correlation (denoted as `spe') 
and common ratio of architectures in groundtruth ranking against the average time in seconds. Especially on CIFAR10 and CIFAR100
large gaps in performance can be seen at high-performing architecture bins.

\paragraph{Finding the training threshold:} We construct a shallow pipeline using Histogram-of-Oriented-Gradients (HoG) 
\citep{dalal2005histograms} as image features and construct a relatively shallow learner by passing the features through
two hidden fully connected layers. This simple pipeline achieves $0.6$ training accuracy on CIFAR10, $0.3$ on CIFAR100 
and $0.2$ accuracy on ImageNet16-120. These numbers were used as the training accuracy threshold for stage 1 of \methodname
with respective datasets.

\paragraph{Random Search with FEAR:} On CIFAR100 and ImageNet16-120 we ran random search (RS) with \methodname (RS-\methodname) 
$10$ times with different random
seeds where the search cut-off any architecture which exceeded $4.0$ times the fastest time to reach threshold training
accuracy encountered so far. Each method got a budget of $500$ architectures. As shown in Table \ref{table:random_search} RS-\methodname 
can get similar final accuracy by being $\approx2.4$ times faster.  See 
Figure \ref{fig:early_rejection_illus} in Appendix \ref{app:details_random_search} for an intuitive explanation of how \methodname
early rejects weaker architectures.

\begin{table}[ht]
    \centering
    \begin{tabular}{lll}
        \toprule
          & \thead{\methodname ($4.0$ * \texttt{fastest}) \\ top1 (\%), duration (s)}  & \thead{\shortreg ($50$ epochs) \\ top1 (\%), duration (s)} \\
        \midrule
        CIFAR10 & $\mathbf{93.97^{\pm0.08}}$, $\mathbf{90560^{\pm845}}$ & $94.10^{\pm0.10*}$, $347643^{\pm2287}$ \\
        CIFAR100 & $\mathbf{72.04^{\pm0.29}}$, $\mathbf{142550^{\pm3106}}$ & $72.08^{\pm0.30}$, $347640^{\pm1674}$ \\
        ImageNet16-120 & $\mathbf{45.97^{\pm0.17}}$, $\mathbf{214824^{\pm4674}}$ & $45.64^{\pm0.21}$, $528454^{\pm4901}$\\
        \bottomrule
    \end{tabular}
    \caption{When used as the architecture
    evaluation scheme in random search, \methodname can achieve same or more top-1 accuracy as \shortreg while
    using $\approx 2.4$ times less search duration on CIFAR100 and ImageNet16-120. On CIFAR10 it takes $\approx 3.8$ times
    less duration. Each experiment was run $10$ times with different random seeds. 
    (*\shortreg on CIFAR10 had finished 7 runs at the time of writing).}
    \label{table:random_search}
\end{table}

\begin{table}[htpb!]
    \begin{subtable}[ht]{\textwidth}
    \footnotesize
    \centering
    \begin{tabular}{c|p{3.0cm}p{3.0cm}|p{3.0cm}p{3.0cm}}
        \toprule
        Top \% & \methodname (spe, s) & Nearest Pareto (spe, s) & \methodname (common, s) & Nearest Pareto (common, s) \\
        \midrule
        $10$ & $\mathbf{0.22}$, $\mathbf{133.22^{\pm 4.09}}$ & $0.16$, $138.28^{\pm 2.25}$ & $\mathbf{0.59}$, $\mathbf{133.22^{\pm 4.09}}$ & $0.40$, $138.28^{\pm 2.25}$ \\
        $20$ & $\mathbf{0.40}$, $\mathbf{133.04^{\pm 2.76}}$ & $0.19$, $134.68^{\pm 1.50}$ & $\mathbf{0.60}$, $\mathbf{133.04^{\pm 2.76}}$ & $0.57$, $134.68^{\pm 1.50}$ \\
        $30$ & $\mathbf{0.41}$, $\mathbf{132.76^{\pm 2.24}}$ & $0.28$, $140.13^{\pm 1.46}$ & $\mathbf{0.70}$, $\mathbf{132.76^{\pm 2.24}}$ & $0.70$, $132.77^{\pm 2.24}$ \\
        $40$ & $\mathbf{0.43}$, $\mathbf{131.41^{\pm 2.00}}$ & $0.34$, $135.31^{\pm 1.32}$ & $\mathbf{0.78}$, $\mathbf{131.41^{\pm 2.00}}$ & $0.72$, $135.31^{\pm 1.32}$ \\
        $50$ & $\mathbf{0.55}$, $\mathbf{134.48^{\pm 1.97}}$ & $0.50$, $136.26^{\pm 1.14}$ & $\mathbf{0.79}$, $\mathbf{134.48^{\pm 1.97}}$ & $0.79$, $136.26^{\pm 1.14}$ \\
        $100$ & $0.83$, $236.36^{\pm 9.39}$ & $\mathbf{0.90}$, $\mathbf{218.97^{\pm 1.44}}$ & $1.0$, $236.36^{\pm 9.39}$ & $\mathbf{1.00}$, $\mathbf{218.97^{\pm 1.44}}$ \\
        \bottomrule
    \end{tabular}
    \label{table:natsbench_cifar10}
    \caption{Natsbench CIFAR 10}
    \end{subtable}
    
    \begin{subtable}[ht]{\textwidth}
    \footnotesize
    \centering
    \begin{tabular}{c|p{3.0cm}p{3.0cm}|p{3.0cm}p{3.0cm}}
        \toprule
        Top \% & \methodname (spe, s) & Nearest Pareto (spe, s) & \methodname (common, s) & Nearest Pareto (common, s) \\
        \midrule
        $10$ & $\mathbf{0.61}$, $\mathbf{213.07^{\pm 6.33}}$ & $0.42$, $264.68^{\pm 3.52}$ & $\mathbf{0.63}$, $\mathbf{213.04^{\pm 6.33}}$ & $0.63$, $264.68^{\pm 3.52}$ \\
        $20$ & $\mathbf{0.51}$, $\mathbf{208.84^{\pm 4.36}}$ & $0.25$, $257.12^{\pm 2.43}$ & $\mathbf{0.79}$, $\mathbf{208.84^{\pm 4.36}}$ & $0.79$, $257.12^{\pm 2.43}$ \\
        $30$ & $\mathbf{0.66}$, $\mathbf{207.19^{\pm 3.63}}$ & $0.15$, $249.86^{\pm 2.03}$ & $\mathbf{0.79}$, $\mathbf{207.19^{\pm 3.63}}$ & $0.79$, $249.86^{\pm 2.03}$ \\
        $40$ & $\mathbf{0.70}$, $\mathbf{205.13^{\pm 3.16}}$ & $0.27$, $253.67^{\pm 2.18}$ & $\mathbf{0.85}$, $\mathbf{205.13^{\pm 3.17}}$ & $0.85$, $244.96^{\pm 1.85}$ \\
        $50$ & $\mathbf{0.76}$, $\mathbf{201.65^{\pm 2.78}}$ & $0.26$, $247.37^{\pm 1.97}$ & $\mathbf{0.90}$, $\mathbf{201.65^{\pm 2.78}}$ & $0.90$, $239.41^{\pm 1.66}$ \\
        $100$ & $\mathbf{0.93}$, $\mathbf{313.52^{\pm 9.31}}$ & $0.90$, $329.79^{\pm 2.19}$ & $1.0$, $313.52^{\pm 9.31}$ & $\mathbf{1.00}$, $\mathbf{281.30^{\pm 2.42}}$ \\
        \bottomrule
    \end{tabular}
    \label{table:natsbench_cifar100}
    \caption{Natsbench CIFAR 100}
    \end{subtable}
    
    \begin{subtable}[ht]{\textwidth}
    \footnotesize
    \centering
    \begin{tabular}{c|p{3.0cm}p{3.0cm}|p{3.0cm}p{3.0cm}}
        \toprule
        Top \% & \methodname (spe, s) & Nearest Pareto (spe, s) & \methodname (common, s) & Nearest Pareto (common, s) \\
        \midrule
        $10$ & $0.56$, $481.75^{\pm 17.14}$ & $\mathbf{0.56}$, $\mathbf{468.17^{\pm 6.07}}$ & $\mathbf{0.76}$, $\mathbf{481.75^{\pm 17.14}}$ & $0.72$, $510.86^{\pm 6.77}$ \\
        $20$ & $\mathbf{0.73}$, $\mathbf{498.91^{\pm 14.52}}$ & $0.64$, $530.86^{\pm 5.98}$ & $0.82$, $498.91^{\pm 14.52}$ & $\mathbf{0.81}$, $\mathbf{445.89^{\pm 5.05}}$ \\
        $30$ & $\mathbf{0.78}$, $\mathbf{486.40^{\pm 11.56}}$ & $0.69$, $514.65^{\pm 4.99}$ & $0.83$, $486.40^{\pm 11.56}$ & $\mathbf{0.85}$, $\mathbf{470.47^{\pm 4.58}}$ \\
        $40$ & $\mathbf{0.79}$, $\mathbf{491.91^{\pm 10.74}}$ & $0.75$, $493.66^{\pm 4.61}$ & $0.90$, $491.91^{\pm 10.74}$ & $0.91$, $493.66^{\pm 4.61}$ \\
        $50$ & $0.84$, $571.37^{\pm 10.73}$ & $\mathbf{0.83}$, $\mathbf{532.04^{\pm 4.70}}$ & $0.91$, $511.37^{\pm 10.73}$ & $0.88$, $510.74^{\pm 5.04}$ \\
        $100$ & $0.95$, $1070.88^{\pm 35.88}$ & $\mathbf{0.96}$, $\mathbf{822.05^{\pm 8.14}}$ & $1.0$, $1070.88^{\pm 35.88}$ & $\mathbf{1.00}$, $\mathbf{822.05^{\pm 8.14}}$ \\
        \bottomrule
    \end{tabular}
    \label{table:natsbench_imagenet16}
    \caption{Natsbench ImageNet16-120}
    \end{subtable}
    \caption{(Left) Spearman's correlation (`spe') comparison between \methodname and the nearest point on the pareto-frontier
    etched out by \shortreg variants with respect to average wall-clock time (s).
    (Right) Ratio of overlap in bins between rankings of \methodname and nearest pareto-frontier point of \shortreg. 
    The nearest pareto-frontier points are found by inspecting figures \ref{fig:cifar10_natsbench}, \ref{fig:cifar100_natsbench} 
    and \ref{fig:imagenet16-120_natsbench}}
    \label{tab:three_spe_cr}
\end{table}

\subsection{Deeper Dive into Zero-Cost Measures}
\label{deep_dive_zerocost}

\begin{table}[htbp!]
    \centering
    \begin{tabular}{ll}
        \toprule
        Method & Spearman Corr. \\
        \midrule
        \synflow & $-0.000437$ \\
        \jacobcov & $-0.13$ \\
        \snip & $-0.31$ \\
        \fisher & $-0.42$ \\
        \grasp & $-0.25$ \\
        \synflowbn & $0.18$ \\
        \methodname & $0.55$ \\
        \bottomrule
    \end{tabular}
    \caption{Zero-Cost measures on \synthetic achieve low correlation while 
    \methodname still gets reasonable performance. This phenomenon is especially not 
    amenable for data-agnostic measures like \synflow which ignore the dataset completely.}
    \label{table:zerocost_synthetic}
\end{table}

As discussed in Section \ref{related_work}, \citet{abdelfattah2021zerocost} propose using 
pruning-at-initialization methods like \synflow, \snip, \grasp, \fisher etc for ranking
architectures without any training by summing up the per-weight saliency scores to come up 
with an overall architecture score. In thorough experiments, \synflow emerged as a 
good ranking measure with a majority voting scheme with \synflow, \jacobcov and \snip emerging
as the best overall. As discussed in Section \ref{related_work},
\synflow['s] good performance is a bit perplexing since it is a data-agnostic
measure. It suggests that there are inherently good and bad architectures and the particulars of 
the dataset should not matter. In order to investigate this we created a synthetic dataset
with properties such that it would be a drop-in for CIFAR10. Specifically we created 
random Gaussian images of dimension \texttt{[32, 32, 3]} and mean $0$ and variance $1$. 
Each image was assigned a class
label in $(0,9)$ by passing each image through $10$ randomly initialized neural networks
and picking the id of the network which assigned the image a maximum score. Each of the 
networks has a simple architecture of a linear layer with dimension $3072$, followed by a 
\texttt{ReLu} layer, followed by a linear layer which produces a single scalar output.
A dataset of $60000$ images was generated with $50000$ training and the rest held-out as a 
test set. Each class has $6000$ examples. We refer to this dataset as \synthetic.

The same set of $1000$ randomly sampled architectures used in above experiments
were evaluated on this dataset using 
\methodname and the various zero-cost measures. Table \ref{table:zerocost_synthetic} shows that the zero-cost
measures have almost no correlation with rankings while \methodname still works reasonably. 
Also note that the rankings via \synflow which ignore the dataset are no longer valid on 
\synthetic. This means that architectures which performed really well on CIFAR10 don't 
work as well on \synthetic. This is at least an existence proof of the fact that performance of an
architecture is also a function of the dataset and task and is not an inherent property only of the
topology of the network. This is also empirically shown recently by \citet{tuggener2021optimize}.

\begin{figure}
  \centering
  \includegraphics[width=\textwidth]{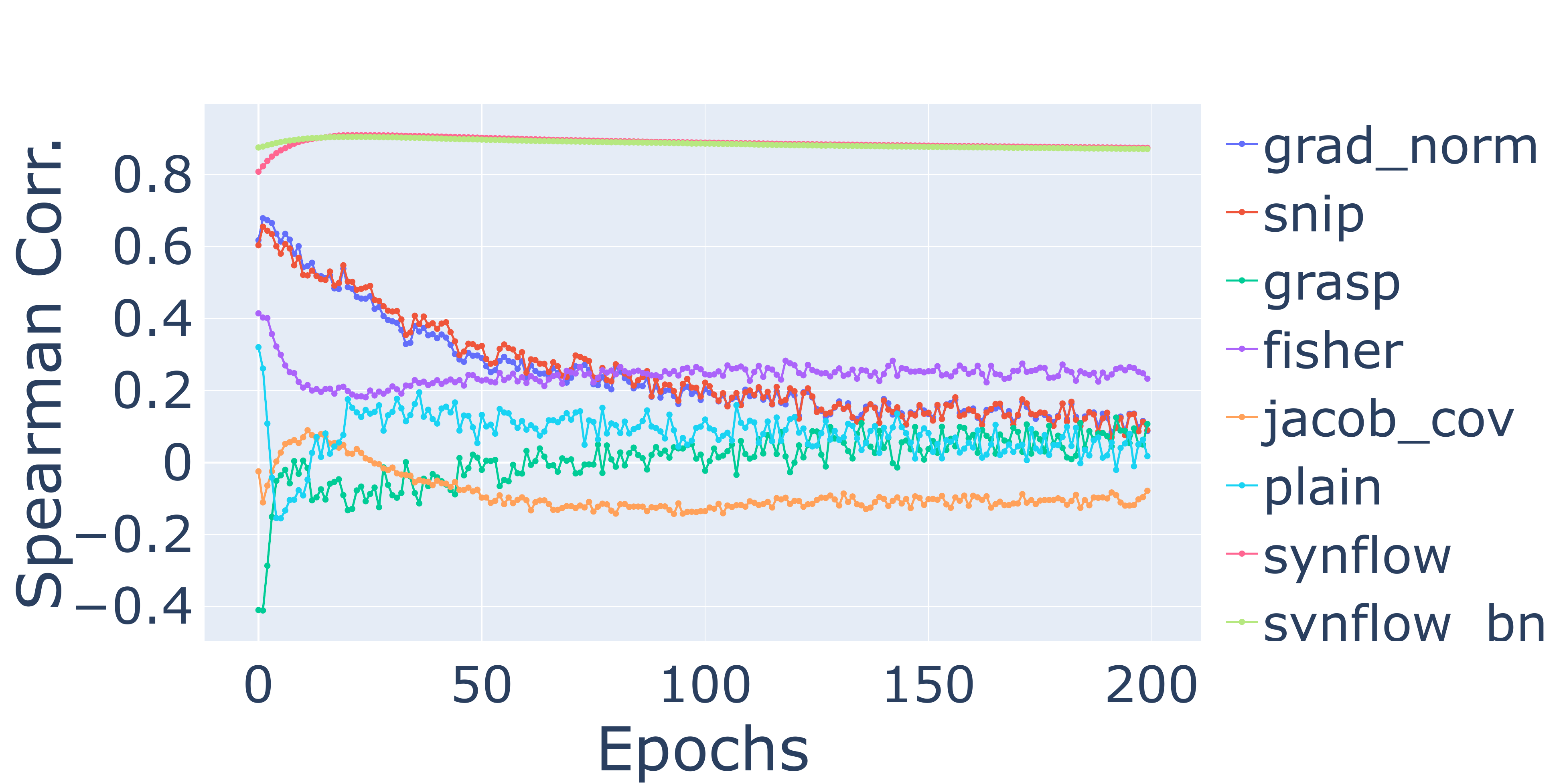}
  \caption{Evaluating the ranking performance of zero-cost measures after each epoch of training. 
  Measures like \grasp and \jacobcov demonstrate large degradation in performance after even a single 
  epoch of training. \snip and \gradnorm decay gradually.}
  \label{fig:zerocost_epochs_c10}
\end{figure}

In Figure \ref{fig:zerocost_epochs_c10} we evaluated zero-cost measures after each epoch of training
on CIFAR10 for the $1000$ randomly sampled architectures from Natsbench topological search space. We
find that measures like \snip and \gradnorm gradually degrade in rank correlation as the network trains. 
\jacobcov and \grasp at initialization have Spearman of $0.69$ and $0.63$ respectively but after even 
one epoch of training drastically degrade to $-0.02$ and $-0.41$. Note from figures \ref{fig:cifar100_natsbench_cropped},
\ref{fig:cifar10_natsbench}, \ref{fig:cifar100_natsbench}, and \ref{fig:imagenet16-120_natsbench} 
that ranking architectures via training error even after one or two epochs of training 
leads to much better ranking correlation.

Note that measures derived from \snip, \grasp, \synflow are intended originally for the task of \emph{pruning}
architectures at initialization. So it is perhaps not surprising that the saliency scores when summed-up
over individual weights to provide a global architecture score doesn't exhibit good ranking performance
as the network is trained.

\section{Conclusion}
We have presented a simple but powerful fast architecture ranking scheme (\methodname) which 
can be used in the inner loop of \emph{any} discrete NAS algorithm to speed-up architecture 
evaluation. We have shown on standard NAS benchmarks that \methodname is effective and in the
inner loop of even random search can drastically speed-up evaluation without loss of accuracy.

In future work we aim to validate \methodname on the state-of-the-art discrete search methods
like Bayesian optimization-based techniques \citep{whiteBananas} or even simpler techniques which have been surprisingly 
beneficial on NAS benchmarks like \citep{white2020local}. With enough compute we will aim to 
validate on larger search spaces like DARTS via Nasbench-301 \citep{siems2020nasbench301} and search spaces 
around Transformer-like architectures \citep{hatWang2020,tsai2020oneshottransformers}.

\bibliographystyle{unsrtnat}
\bibliography{references}

\clearpage

\appendix

\section{Appendix}
\label{appendix}

\subsection{Reproducibility and Best Practices Checklist in NAS}
\label{app:naschecklist}

We use the best practices by \citep{nasChecklist} to foster reproducibility and 
do better empirical NAS research. 

\begin{enumerate}
    \item Best practices for releasing code.
    \begin{enumerate}
        \item Code for the training pipeline used to evaluate the final architectures. - \textcolor{blue}{Yes.}
        \item Code for the search space. - \textcolor{blue}{Yes.}
        \item The hyperparameters used for the final evaluation pipeline as well as random seeds. - \textcolor{blue}{Yes.}
        \item Code for your NAS method. - \textcolor{blue}{Yes.}
        \item Hyperparameters for your NAS method, as well as random seeds. - \textcolor{blue}{Yes.}
    \end{enumerate}
    \item Best practices for comparing NAS methods.
    \begin{enumerate}
        \item For all NAS methods you compare, did you use exactly the same NAS benchmark, 
        including the same dataset (with the same training-test split), search space and code for training the architectures and hyperparameters for that code? - \textcolor{blue}{Yes.}
        \item Did you control for confounding factors (different hardware, versions of DL libraries, different runtimes for the different methods)? - \textcolor{blue}{Yes. Specifically we run all the baselines ourselves on the same hardware.}
        \item Did you run ablation studies? - \textcolor{blue}{Yes.}
        \item Did you use the same evaluation protocol for the methods being compared? - \textcolor{blue}{Yes.}
        \item Did you compare performance over time? - \textcolor{blue}{Yes.}
        \item Did you compare to random search? - \textcolor{blue}{Yes.}
        \item Did you perform multiple runs of your experiments and report seeds? - \textcolor{blue}{Yes.}
        \item Did you use tabular or surrogate benchmarks for in-depth evaluations? - \textcolor{blue}{Yes.}
    \end{enumerate}
    \item Best practices for reporting important details. - \textcolor{blue}{Yes}
    \begin{enumerate}
        \item Did you report how you tuned hyperparameters, and what time and resources this required? - \textcolor{blue}{Yes.}
        \item Did you report the time for the entire end-to-end NAS method (rather than, e.g., only for the search phase)? - \textcolor{blue}{Yes.}
        \item Did you report all the details of your experimental setup? - \textcolor{blue}{Yes.}
    \end{enumerate}

\end{enumerate}

\subsection{Detailed Random Search Experiments}
\label{app:details_random_search}

\begin{algorithm}[h]
\caption{Random Search with \methodname}
\label{alg:random_search_fear}
\begin{algorithmic}
\STATE Input: Architecture search space $\mathcal{A}$, ratio of fastest time to reach threshold $r$, 
maximum budget of architectures: $B$, threshold accuracy to train to: $t$
\STATE Output: Best architecture $a \in \mathcal{A}$
\STATE $\mathrm{fastest\_till\_now} = \infty$
\STATE $\mathrm{best\_train\_till\_now} = -\infty$
\STATE $\mathrm{best\_arch} = \mathrm{None}$
\FOR{$i=1\ \mathrm{to}\ B$}
\STATE $a = \mathrm{random\_sample}(A)$ \algorithmiccomment{\textcolor{cyan}{Uniformly random sample an architecture.}}
\STATE $\mathrm{time\_allowed} = r \times \mathrm{fastest\_till\_now}$ \algorithmiccomment{\textcolor{cyan}{Time allowed for architecture evaluation by \methodname}.}
\STATE $\mathrm{result}, \mathrm{time\_taken} = \mathrm{FEAR}(a, t, \mathrm{fastest\_till\_now})$ \algorithmiccomment{\textcolor{cyan}{\methodname evaluate architecture.}}
\IF{$\mathrm{not}\ \mathrm{result}$} 
    \STATE $\mathrm{continue}$ \algorithmiccomment{\textcolor{cyan}{If \methodname early-rejected then move on to next architecture.}}
\ELSE
    \IF{$\mathrm{result} > \mathrm{best\_train\_till\_now}$}
        \STATE $\mathrm{best\_train\_till\_now} = \mathrm{result}$ \algorithmiccomment{\textcolor{cyan}{If a better architecture was found then update.}}
        \STATE $\mathrm{best\_arch} = a$ 
        \IF{$\mathrm{time\_taken} < \mathrm{fastest\_till\_now}$}
            \STATE $\mathrm{fastest\_till\_now} = \mathrm{time\_taken}$ \algorithmiccomment{\textcolor{cyan}{Update fastest time if faster than current fastest.}}
        \ENDIF
    \ENDIF
\ENDIF
\ENDFOR
\RETURN $\mathrm{best\_arch}$
\end{algorithmic}
\end{algorithm}

\begin{figure}[htbp!]
  \centering
  \includegraphics[width=\textwidth]{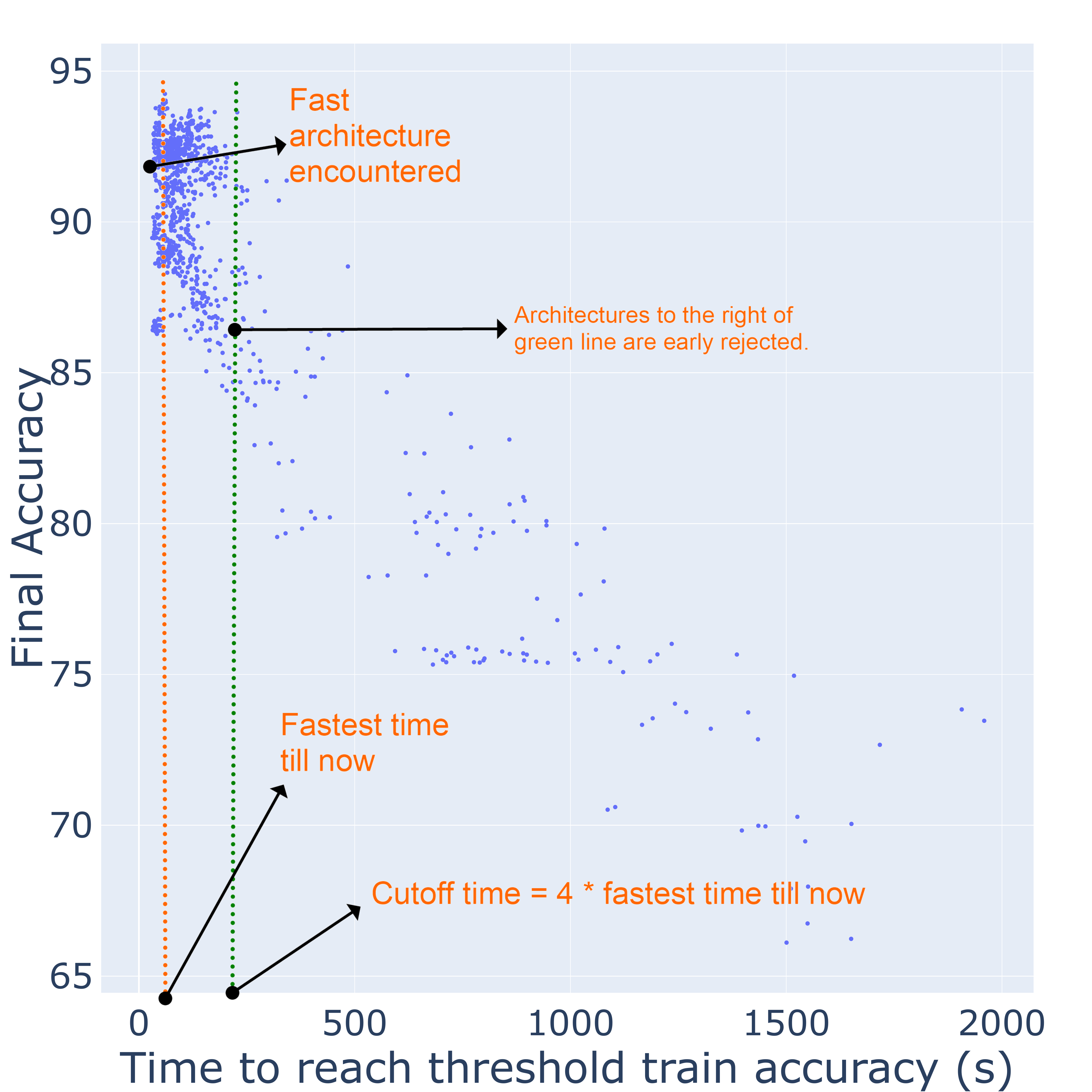}
  \caption{In this illustration on $1000$ architectures from Natsbench topology search space on CIFAR10
  we show how one can use FEAR in the inner loop of any discrete search algorithm to 
  early-reject many architectures thus saving time without losing final accuracy. For simplicity we consider
  random search here. A simple modification to random search is made where the fastest time for any architecture to 
  reach threshold accuracy is tracked. If any subsequent architecture improves on this time then it is updated. If 
  any subsequent architecture during evaluation takes more than \texttt{x $\times$ fastest\_till\_now} (where 
  \texttt{x} is usually $4.0$ in our experiments as that leads to retaining most of the top ones) then evaluation
  is terminated and control moves on to the next randomly sampled architecture. Such architectures which are early 
  rejected from evaluation are 
  represented by the region to the right of the green vertical dotted line in the figure. As more and more 
  architectures are evaluated the estimate of the \texttt{fastest\_till\_now} improves and so does early termination of 
  weaker architectures. And this benefit of early stopping weaker architectures is evident in our experiments 
  where we see that one can reduce the search time by $\approx2.4$ times on 
  CIFAR100 and ImageNet16-120 without losing accuracy as detailed in Section \ref{experiments}.}
  \label{fig:early_rejection_illus}
\end{figure}

Algorithm \ref{alg:random_search_fear} shows the simple modifications made to vanilla random search to enable
architecture evaluation with \methodname and early-reject weaker architectures which inevitably take much longer
to reach threshold accuracy than stronger candidates. See Figure \ref{fig:early_rejection_illus} and associated
caption for an intuitive illustration of how one can early reject most architectures from the search space.

\clearpage

\subsection{Detailed Ranking Plots on Natsbench Topological Search Space}
\label{app:detailed_ranking}

\begin{figure}[htbp!]
  \centering
  \includegraphics[width=\textwidth]{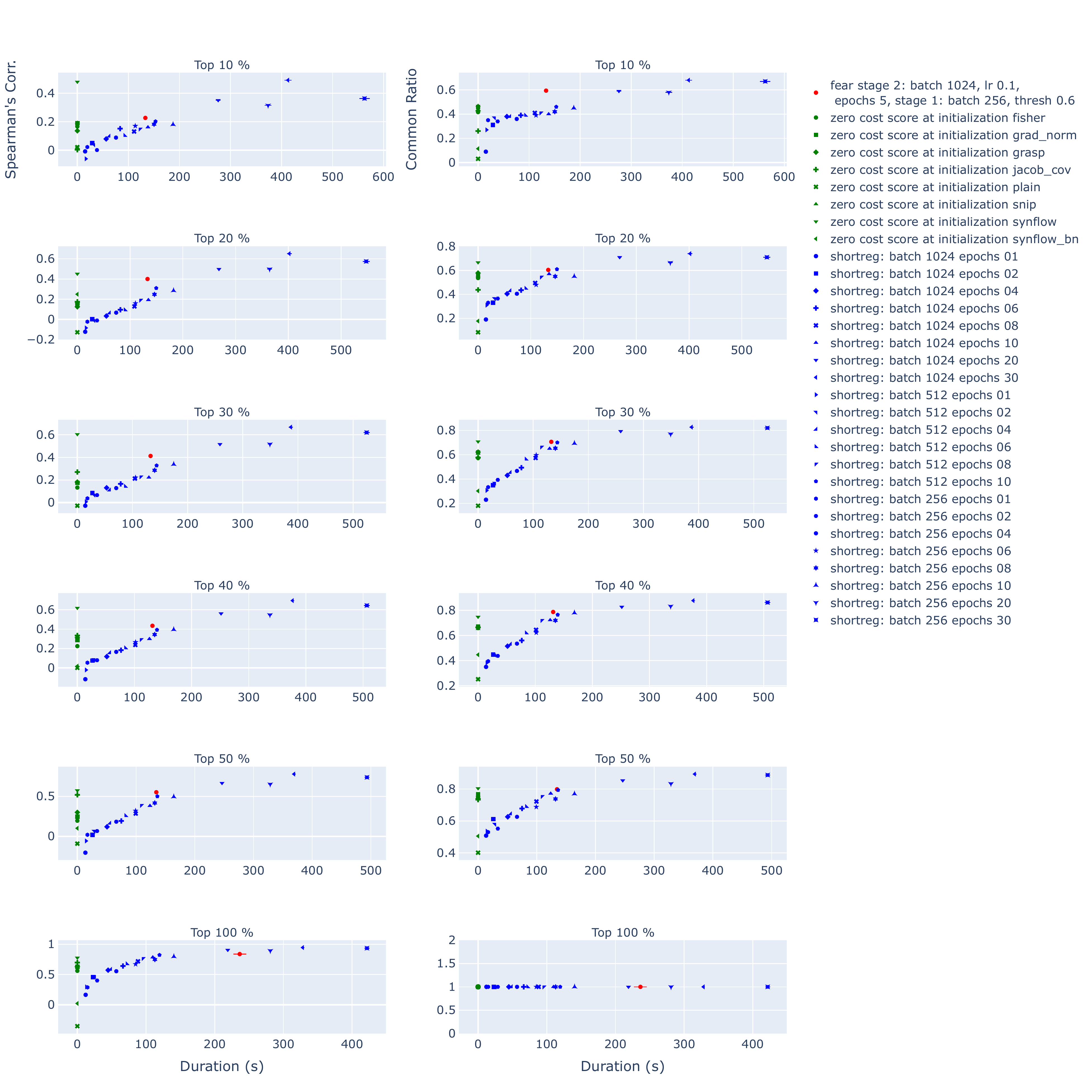}
  \caption{[Left] Average duration vs. Spearman's correlation and [Right] averge duration vs. common ratio over the top $x\%$ of 
  the $1000$ architectures sampled from Natsbench topological search space on CIFAR10. We also show the various zero-cost measures
  from \citet{abdelfattah2021zerocost} in green.}
  \label{fig:cifar10_natsbench}
\end{figure}

\begin{figure}[htbp!]
  \centering
  \includegraphics[width=\textwidth]{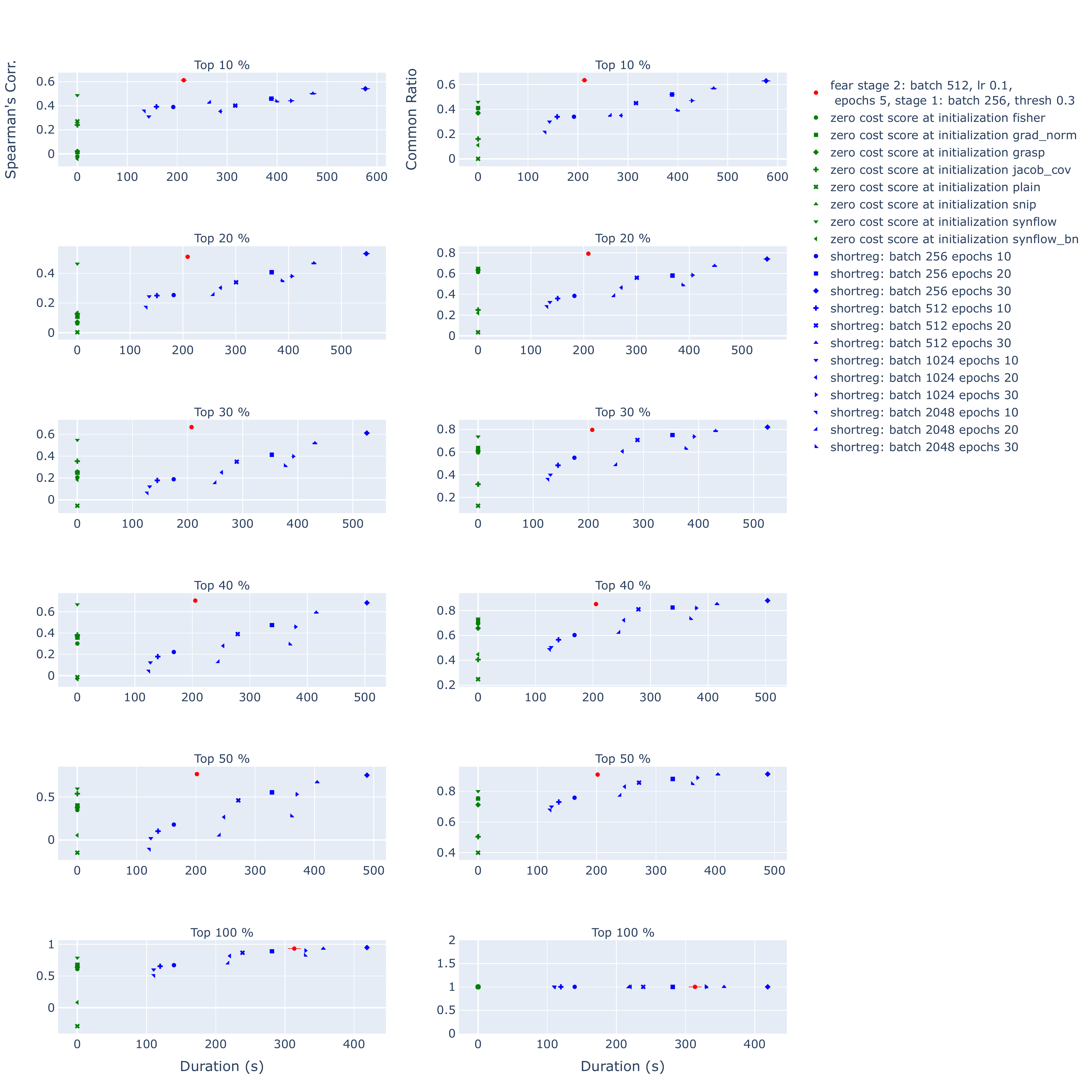}
  \caption{[Left] Average duration vs. Spearman's correlation and [Right] average duration vs. common ratio over the top $x\%$ of 
  the $1000$ architectures sampled from Natsbench topological search space on CIFAR100. We also show the various zero-cost measures
  from \citet{abdelfattah2021zerocost} in green.}
  \label{fig:cifar100_natsbench}
\end{figure}

\begin{figure}[htbp!]
  \centering
  \includegraphics[width=\textwidth]{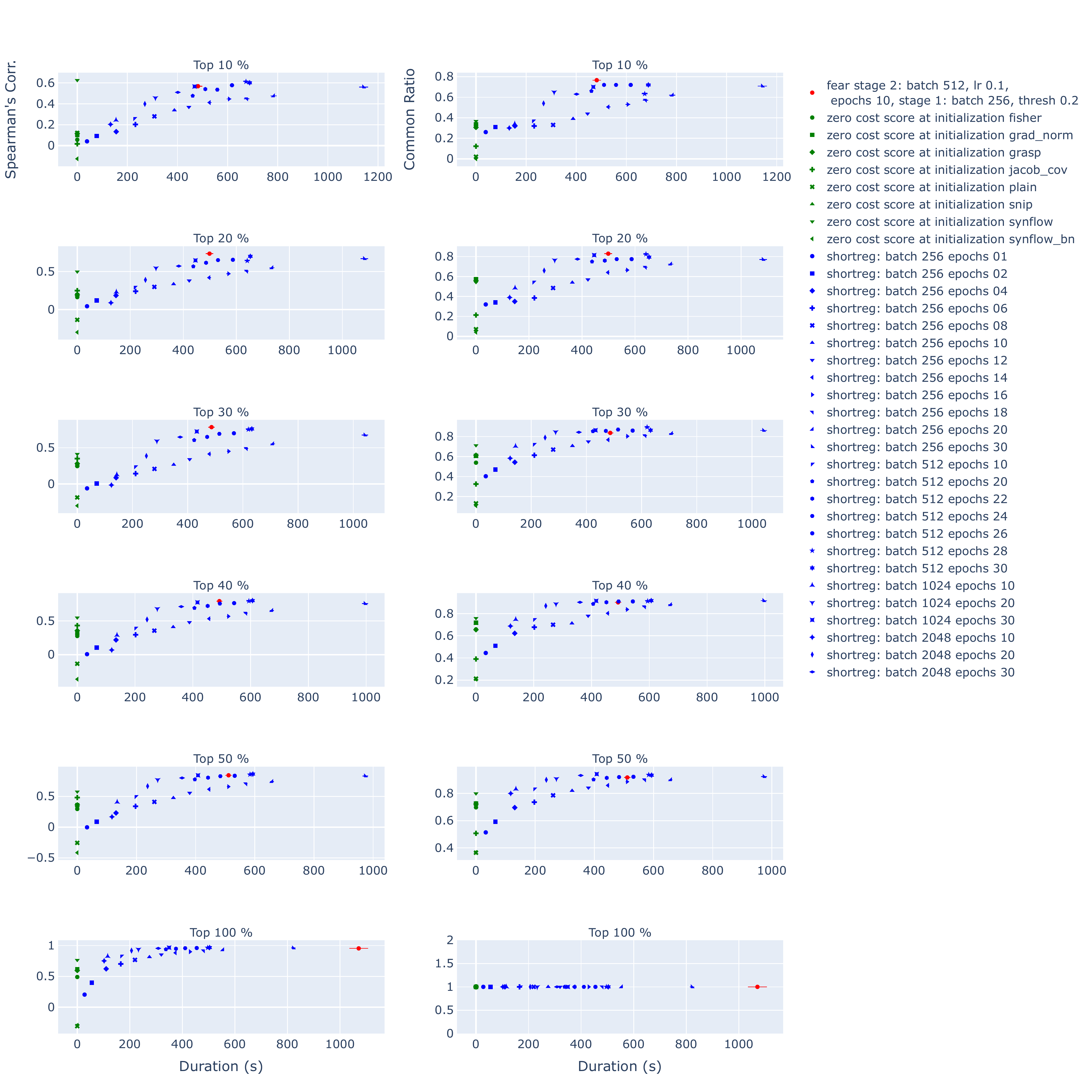}
  \caption{[Left] Average duration vs. Spearman's correlation and [Right] average duration vs. common ratio over the top $x\%$ of 
  the $1000$ architectures sampled from Natsbench topological search space on ImageNet16-120. We also show the various zero-cost measures
  from \citet{abdelfattah2021zerocost} in green.}
  \label{fig:imagenet16-120_natsbench}
\end{figure}

\clearpage

\end{document}